\documentclass[runningheads]{llncs}

\usepackage[T1]{fontenc}
\usepackage{graphicx}
\usepackage{hyperref}
\usepackage{color}

\usepackage{todonotes}
\usepackage{amssymb}
\usepackage{booktabs}
\usepackage{xcolor}
\usepackage{pifont}
\usepackage{cite}
\usepackage{multirow}
\usepackage{support-caption}
\usepackage{subcaption}
\usepackage{makecell}
\usepackage{tabularx}
\usepackage{adjustbox}
\usepackage{nicefrac}

\newcommand{\cmark}{{\color{green!40!black}{\ding{51}}}}
\newcommand{\xmark}{{\color{red}{\ding{55}}}}

\newcommand{\bod}{\mathfrak{B}}
\newcommand{\rec}{\mathfrak{R}}
\newcommand{\sym}{\mathfrak{S}}

\newcommand{\scalar}{[\texttt{S}]}
\newcommand{\binary}{[\texttt{B}]}

\begin{document}
\title{The Power of Properties: Uncovering the Influential Factors in Emotion Classification}
\titlerunning{Influential Factors in Emotion Classification}

\author{
Tim Büchner$^*$\inst{1}\orcidID{0000-0002-6879-552X} \and
Niklas Penzel$^*$\inst{1}\orcidID{0000-0001-8002-4130} \and
Orlando Guntinas-Lichius\inst{2}\orcidID{0000-0001-9671-0784} \and
Joachim Denzler\inst{1}\orcidID{0000-0002-3193-3300}
}

\authorrunning{Büchner, T. et al.}
\institute{
    Computer Vision Group, Friedrich Schiller University Jena, 07743 Jena, Germany
    \email{\{firstname.surname\}@uni-jena.de}
    \and
    Dept. of Otorhinolaryngology, Jena University Hospital, 07747 Jena, Germany
    \\
    \email{orlando.guntinas@uni-jena.de}
}

\maketitle

\begin{NoHyper}
\def\thefootnote{*}\footnotetext{These authors contributed equally to this work.}\def\thefootnote{\arabic{footnote}}
\end{NoHyper}

\begin{abstract}
Facial expression-based human emotion recognition is a critical research area in psychology and medicine.
State-of-the-art classification performance is only reached by end-to-end trained neural networks.
Nevertheless, such black-box models lack transparency in their decision-making processes, prompting efforts to ascertain the rules that underlie classifiers' decisions.
Analyzing single inputs alone fails to expose systematic learned biases.
These biases can be characterized as facial \emph{properties} summarizing abstract information like age or medical conditions.
Therefore, understanding a model's prediction behavior requires an analysis rooted in causality along such selected \emph{properties}.
We demonstrate that up to $91.25\%$ of classifier output behavior changes are statistically significant concerning basic \emph{properties}.
Among those are age, gender, and facial symmetry.
Furthermore, the medical usage of surface electromyography significantly influences emotion prediction.
We introduce a workflow to evaluate explicit \emph{properties} and their impact.
These insights might help medical professionals select and apply classifiers regarding their specialized data and \emph{properties}.

\end{abstract}

\section{Introduction}
Human emotion recognition via facial expressions is essential in psychology and medical research.
High classification performance is achieved by neural networks trained end-to-end and tailored for broad usage on large datasets.
The application of these black box models to unseen data often leads to irregular behavior.
Previous research, for example, \cite{yang2023change}, finds a loss in predictive performance when observing various so-called subpopulation shifts.
While we follow a similar approach, the performance decline attributed to subpopulation shifts is insufficient.
To uncover potential causes, we model these shifts as changes in what we call \emph{properties}.
Consequently, by investigating different \emph{property} manifestations, we can uncover deviations in the model behavior. 
Hence, we go beyond simple predictive performance and uncover influential factors in emotion classification.

Some of these potential factors (\emph{properties}) such as age or gender, have a broad impact on the visual appearance of a face.
Regarding emotion recognition, this is even more evident in the case of medical conditions influencing mimicry.
Particularly in facial palsy cases, where there is unilateral paralysis of the facial nerve.
The pronounced facial asymmetry may impact a model's prediction or muscle activation studies with joint surface electromyography (sEMG)~\cite{guntinas2023high,buechner2023lets,buechner2023improved}.

We assess HSEmotion-7~\cite{savchenko2023facial} and ResidualMaskNet~\cite{pham2021facial}, specifically their application on medically acquired facial data.
Both models can predict the six basic emotions after Ekman~\cite{ekman1992argument}, with an additional class for neutral expressions.
We record 36 healthy probands to investigate the presence of sEMG electrodes and additionally their artificial removal~\cite{buechner2023improved,buechner2023lets}.
Further, we capture 36 patients with facial palsy to evaluate the influence of facial asymmetry~\cite{buchnerFacesVolumesMeasuring2023}.
This setup enables us to capture a multitude of relevant \emph{properties}.
Allowing us to extend a general performance analysis for subpopulation shifts~\cite{yang2023change} by utilizing additional explanations~\cite{reimers2020determining}.
These explanations are based on causal principles~\cite{reimers2020determining}, permitting us to test for statistically significant deviations in model behavior.

Our experimental results reveal two principal insights.
First, both models exhibit a variation in predictive performance;
Second, and more critically, there is a significant ($p < 0.01$) behavior change concerning the manifestations of varying \emph{properties}.
While for some, the shift in model behavior only significantly occurs for some emotions, other \emph{properties} are important irrespective of the predicted class.
Examples include visibly attached sEMG electrodes and whether a person suffers from unilateral facial palsy.
We find statistically significant changes in the model behavior in up to $91.25\%$ of the analyzed \emph{properties}.

\section{Facial Emotion Classification}
Facial emotion classification is a broad research field with many possible applications, especially in psychology and medicine.
State-of-the-art performance is currently attainable exclusively through end-to-end trained convolutional neural networks.
Our study focuses on two such models: HSEmotion-7 (HSE-7)~\cite{savchenko2023facial} and ResidualMaskNet (RMN)~\cite{pham2021facial}.
Both models predict the six basic emotions defined by Ekman~\cite{ekman1992argument}, with an added class for neutral expressions.

A general reduction in prediction accuracy is anticipated~\cite{yang2023change}.
However, our interests are the subpopulation shifts resulting from different \emph{property} manifestations.
Therefore, we do not fine-tune to prevent distorting the interpretability of the general behavior analysis.
Toward this goal, we require a custom evaluation dataset that captures \emph{properties} typically unaccounted for in large datasets.

\begin{table}[t]
    \centering
    \caption{
        We list the accuracies by emotion, sEMG attachment, and model (without fine-tuning) for probands and patients.
        A reference image is shown for each set.
        The patients are recorded only for the \emph{happy} expression without sEMG.
    }
    \newcolumntype{C}{>{\centering}p{1.0cm}}
\newcolumntype{R}{>{\raggedleft\arraybackslash}p{1.0cm}}
\newcolumntype{L}{>{\raggedright\arraybackslash}p{1.0cm}}
\setlength{\tabcolsep}{3pt}

\begin{tabularx}{\textwidth}{XRLRLRLRL}
\toprule
            & \multicolumn{6}{c}{Probands} & \multicolumn{2}{c}{Patients} \\
            \cline{2-7}\cline{8-9}
\multicolumn{1}{r}{sEMG}        & \multicolumn{2}{c}{Without} 
            & \multicolumn{2}{c}{Attached} 
            & \multicolumn{2}{c}{Removed} 
            & \multicolumn{2}{c}{Without} \\
\midrule
Emotion     & RMN   & HSE-7 & RMN   & HSE-7 & RMN   & HSE-7 & RMN   & HSE-7  \\
\midrule
angry       & 82.29 & 66.32 & 36.25 & 36.07 & 82.86 & 54.82 & -     & -     \\
disgusted   & 60.42 & 82.99 &  0.00 & 10.89 & 44.11 & 66.07 & -     & -     \\
fearful     & 33.33 & 60.76 &  0.71 & 58.39 & 20.36 & 45.89 & -     & -     \\
happy       & 95.83 & 87.85 & 66.96 & 45.89 & 73.04 & 46.07 & 60.64 & 65.81 \\
sad         & 13.19 & 81.25 &  0.18 & 75.18 &  8.57 & 76.96 & -     & -     \\
surprised   & 65.62 & 55.90 & 98.21 & 45.00 & 47.50 & 49.46 & -     & -     \\
Mean Acc.   & 58.45 & 72.51 & 33.72 & 45.24 & 46.07 & 56.54 & -& - \\
\midrule
\makecell{Reference \\Image} 
            & \multicolumn{2}{m{2.1cm}}{
            \centering
            \includegraphics[width=0.10\textwidth]{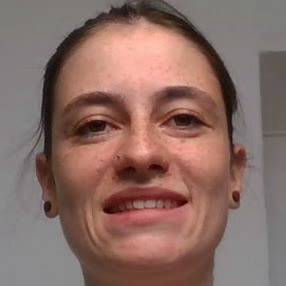}}
            & \multicolumn{2}{m{2cm}}{
            \centering
            \includegraphics[width=0.10\textwidth]{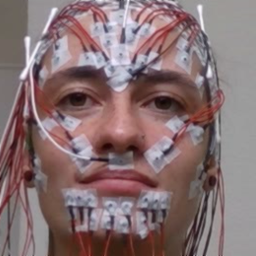}}
            & \multicolumn{2}{m{2cm}}{
            \centering
            \includegraphics[width=0.10\textwidth]{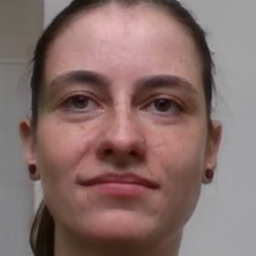}}
            & \multicolumn{2}{m{2.1cm}}{
            \centering
            \includegraphics[width=0.093\textwidth]{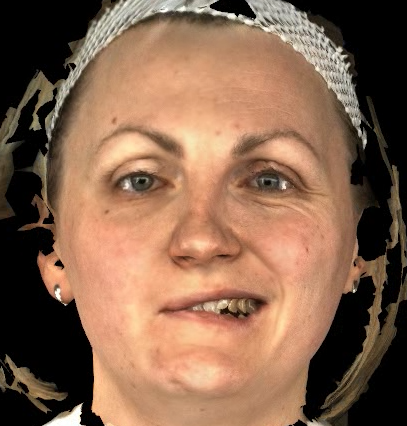}}\\
\bottomrule
\end{tabularx}

    \label{tab:accuracy}
\end{table}

\begin{table}[t]
    \caption{
        We list our selected \emph{properties} used to examine model behavior.
        Their manifestations are scalars \scalar~or a binary \binary~value.
    }
    \centering
    \setlength{\tabcolsep}{3pt}

\begin{tabular}{crcl}
\toprule
& Property & Abbr. & Description \\ 
\midrule
\multirow{4}{*}{\rotatebox[origin=c]{90}{Bodily}} 
& Age \scalar              & $\bod_A$      & The participant's age at the moment of recording \\
& Weight \scalar           & $\bod_W$      & The measured weight at the moment of recording \\
& Gender \binary           & $\bod_G$      & The self-declared gender during questionnaire \\
& Facial Palsy \binary     & $\bod_F$      & The participant suffers from unilateral facial palsy \\
\hline
\multirow{4}{*}{\rotatebox[origin=c]{90}{Recording}} 
& Participant ID \scalar   & $\rec_P$      & Unique participant ID, to check individuum bias\\
& Session ID \scalar       & $\rec_S$      & Recording session, to check for task memorization \\
& Attached sEMG \binary    & $\rec_A$      & If sEMG electrodes are attached~\cite{guntinas2023high} \\
& Removed sEMG \binary     & $\rec_R$      & If sEMG electrodes are removed artifically~\cite{buechner2023lets,buechner2023improved} \\
\hline
\multirow{4}{*}{\rotatebox[origin=c]{90}{Symmetry}} 
& Facial Volume  \scalar   & $\sym_V$ & The lateral facial volume difference~\cite{buchnerFacesVolumesMeasuring2023} \\
& eye-level dev. \scalar   & $\sym_E$ & The deviation angle for the horizontal eye line~\cite{weiAssessingFacialSymmetry2022}  \\
& midline dev.  \scalar    & $\sym_M$ & The deviation angle for nose bridge line~\cite{weiAssessingFacialSymmetry2022}\\
& LPIPS         \scalar    & $\sym_L$ & The image similarity between facial halves~\cite{zhangUnreasonableEffectivenessDeep2018}\\
\bottomrule
\end{tabular}
    \label{tab:features}
\end{table}

\subsection{Emotion Evaluation Dataset}
Our facial emotion images were captured using a standardized procedure where participants had to mimic the six basic emotions four times in a random sequence.
Therefore, we eliminate the risk of human-annotation bias, depending exclusively on the participants' capacity for facial mimicry.
Using a frontal camera, we collected data from 36 healthy probands (18-67 years, 17 male, 19 female) four times with and twice without attached high-resolution surface electromyography (HR-sEMG) \cite{guntinas2023high}.
We follow the work of~\cite{buechner2023lets,buechner2023improved} to remove sEMG electrodes artificially.
For each of these groups, we list the accuracy per model and emotion in \autoref{tab:accuracy}.
Furthermore, 36 patients (25-72 years, 8 male, 28 female) with unilateral chronic synkinetic facial palsy, which is presumed to be a significant factor affecting classification, were recorded three times using the 3dMD face system (3dMD LLC, Georgia, USA).
This system creates 3D scans of the patients.
We simulate a frontal view by calculating the camera position based on facial landmarks~\cite{buchnerFacesVolumesMeasuring2023}, ensuring consistency among participants; see \autoref{tab:accuracy}.
The first set of recordings only includes \emph{happy} expressions.
Combined, our dataset consists of 8,952 annotated images for evaluation.

\subsection{Facial Properties: Selection and Manifestations}\label{sec:foi}
To evaluate the model's behavior regarding subpopulation shifts, we must select \emph{properties}.
The selection criteria are discussed herein, while \autoref{tab:features} provides an overview comprising the manifestation types: scalar \scalar~or binary \binary.
The medical setup grants access to \texttt{bodily} \emph{properties} typically unaccounted for in large datasets.
These include age, weight, gender, and presence of facial palsy.

Additionally, we utilize the experiment repetitions to evaluate model behavior across \texttt{recordings} for the same participant.
Individual recording sessions are also scrutinized to detect any learned expressions.
We indicate sEMG attachment and its subsequent artificial removal for the probands.
We selected four computed metrics to assess facial \texttt{symmetry}, as facial palsy alone is insufficient for impact evaluation.
We compute the lateral volume difference using the 3D patient scans~\cite{buchnerFacesVolumesMeasuring2023}.
Further, two \emph{properties} leverage facial landmark symmetry~\cite{weiAssessingFacialSymmetry2022}.
LPISP calculates the image similarity between facial halves, with images aligned and center-cropped along the eye-line to reduce rotation artifacts~\cite{zhangUnreasonableEffectivenessDeep2018,buechner2023improved}.

It should be noted that a \emph{property} like \texttt{age} encompasses a complex mixture of features, including wrinkles, hair color, or age spots.
This type of subdivision can be accomplished for nearly all chosen \emph{properties}.
While a more comprehensive analysis would give more insight, it is not feasible within the scope of this study.

\begin{table}[t]
    \centering
    \caption{
        Significant changes ($p < 0.01$) per model and emotion are denoted with \cmark.
        Additionally, we denote the average usage ($\varnothing$) by \emph{property} and model.
    }
    \setlength{\tabcolsep}{2.5pt}
\begin{tabularx}{\textwidth}{l>{\raggedleft\arraybackslash}p{1.6cm}cccc@{\hskip 0.5cm}cccc@{\hskip 0.5cm}ccccc}
    \toprule
    &           & \multicolumn{4}{l}{\hspace{0.8cm}Bodily} & \multicolumn{4}{l}{\hspace{0.36cm}Recording} & \multicolumn{4}{l}{\hspace{0.36cm}Symmetry} & \\
    &           & $\bod_A$ & $\bod_W$ & $\bod_G$ & $\bod_F$ & $\rec_P$  & $\rec_S$ & $\rec_A$ & $\rec_R$ & $\sym_V$ & $\sym_E$ & $\sym_M$ & $\sym_L$ & \\
    \cline{1-15}
    \multirow{8}{*}{\rotatebox[origin=c]{90}{HSEmotion-7 \cite{savchenko2023facial}}} 
    & angry     & \cmark   & \cmark   & \cmark   & \cmark   & \cmark    & \xmark   & \cmark   & \cmark   & \cmark   & \xmark   & \xmark   & \cmark   &\\
    & disgusted & \cmark   & \cmark   & \cmark   & \cmark   & \cmark    & \xmark   & \cmark   & \cmark   & \cmark   & \cmark   & \cmark   & \cmark   &\\
    & fearful   & \cmark   & \cmark   & \cmark   & \cmark   & \cmark    & \cmark   & \cmark   & \cmark   & \cmark   & \cmark   & \xmark   & \cmark   &\\
    & happy     & \cmark   & \cmark   & \xmark   & \cmark   & \xmark    & \xmark   & \cmark   & \cmark   & \cmark   & \cmark   & \xmark   & \cmark   &\\
    & sad       & \cmark   & \cmark   & \cmark   & \cmark   & \cmark    & \cmark   & \cmark   & \cmark   & \xmark   & \cmark   & \xmark   & \cmark   &\\
    & surprised & \xmark   & \cmark   & \cmark   & \cmark   & \cmark    & \xmark   & \cmark   & \cmark   & \cmark   & \xmark   & \cmark   & \cmark   &\\
    & neutral   & \cmark   & \cmark   & \cmark   & \cmark   & \cmark    & \cmark   & \cmark   & \cmark   & \cmark   & \cmark   & \cmark   & \cmark   &\\
    \cline{3-14}
    &  $\varnothing$ 
    & \nicefrac{6}{7} & \nicefrac{7}{7} & \nicefrac{6}{7} & \nicefrac{7}{7} 
    & \nicefrac{6}{7} & \nicefrac{3}{7} & \nicefrac{7}{7} & \nicefrac{7}{7} 
    & \nicefrac{6}{7} & \nicefrac{5}{7} & \nicefrac{3}{7} & \nicefrac{7}{7}
    & $\sum$ \nicefrac{70}{84}
    \\
    \cline{1-15}
    \multirow{8}{*}{\rotatebox[origin=c]{90}{ResMaskNet \cite{pham2021facial}}} 
    & angry     & \cmark   & \cmark   & \xmark   & \cmark   & \cmark    & \cmark   & \cmark   & \cmark   & \cmark   & \xmark   & \cmark   & \cmark   &\\
    & disgusted & \cmark   & \cmark   & \xmark   & \cmark   & \cmark    & \cmark   & \cmark   & \cmark   & \cmark   & \xmark   & \cmark   & \cmark   &\\
    & fearful   & \cmark   & \cmark   & \cmark   & \cmark   & \cmark    & \cmark   & \cmark   & \cmark   & \xmark   & \xmark   & \cmark   & \cmark   &\\
    & happy     & \cmark   & \cmark   & \cmark   & \cmark   & \cmark    & \cmark   & \cmark   & \cmark   & \cmark   & \cmark   & \cmark   & \cmark   &\\
    & sad       & \cmark   & \cmark   & \cmark   & \cmark   & \cmark    & \xmark   & \cmark   & \cmark   & \cmark   & \xmark   & \xmark   & \cmark   &\\
    & surprised & \cmark   & \cmark   & \cmark   & \cmark   & \xmark    & \cmark   & \cmark   & \cmark   & \cmark   & \cmark   & \cmark   & \cmark   &\\
    & neutral   & \cmark   & \cmark   & \cmark   & \cmark   & \cmark    & \cmark   & \cmark   & \cmark   & \xmark   & \cmark   & \cmark   & \cmark   &\\
    \cline{3-14}
    &  $\varnothing$ 
    & \nicefrac{7}{7} & \nicefrac{7}{7} & \nicefrac{5}{7} & \nicefrac{7}{7} 
    & \nicefrac{6}{7} & \nicefrac{6}{7} & \nicefrac{7}{7} & \nicefrac{7}{7}
    & \nicefrac{5}{7} & \nicefrac{3}{7} & \nicefrac{6}{7} & \nicefrac{7}{7}
    & $\sum$ \nicefrac{73}{84}
    \\
    \bottomrule
\end{tabularx}
    \label{tab:ci-results}
\end{table}

\section{Methodology}\label{sec:analysis_method}
Interested in emotion recognition model behavior relative to subpopulation shifts, we base the analysis on selected \emph{properties} and related manifestations.
Following the approach of~\cite{yang2023change}, we assess accuracy change with respect to these shifts, see \autoref{tab:accuracy}.
We build our main analysis upon a causally-grounded method detailed in~\cite{reimers2020determining} to detect statistically significant behavioral shifts in the model in response to property manifestation.
In~\cite{reimers2020determining}, the authors develop a structural causal model (SCM) encompassing supervised learning building on the causality framework of Pearl~\cite{pearl2009causality}.
Using this SCM together with Reichenbach's common cause principle~\cite{reichenbach1956direction}, the question of whether a trained classifier uses a \emph{property} becomes a statistical conditional independence (CI) test~\cite{reimers2020determining}.
The selection of suitable CI tests is a vital hyperparameter choice.
However, in~\cite{shah2020hardness}, the authors prove that there cannot be a non-parametric CI test that controls for Type-I errors (false positives) in all cases.
Following the analysis in~\cite{reimers2021conditional,penzel2022investigating,penzel2023analyzing}, we form a committee of nonlinear tests to assess a model's feature usage~\cite{penzel2022investigating}, specifically conditional HSIC~\cite{fukumizu2007kernel}, RCoT~\cite{strobl2019approximate}, and CMIknn~\cite{runge2018conditional}.
We report the consensus results with a significance level of $p<0.01$ in \autoref{tab:ci-results}.

Since both models output logits use softmax~\cite{savchenko2023facial,pham2021facial}, observable changes in one logit may be counterbalanced by opposing logit groups.
Hence, we verify each emotion to address this, as displayed in \autoref{tab:ci-results}.
This approach enables more profound insight into \emph{property} utilization beyond simple summary statistics.

\section{Results}
\begin{figure}[t]
    \centering
    \begin{subfigure}{0.32\textwidth}
        \centering
        \includegraphics[width=\textwidth, clip, trim={0cm 1.41cm 0cm 1.1cm}]{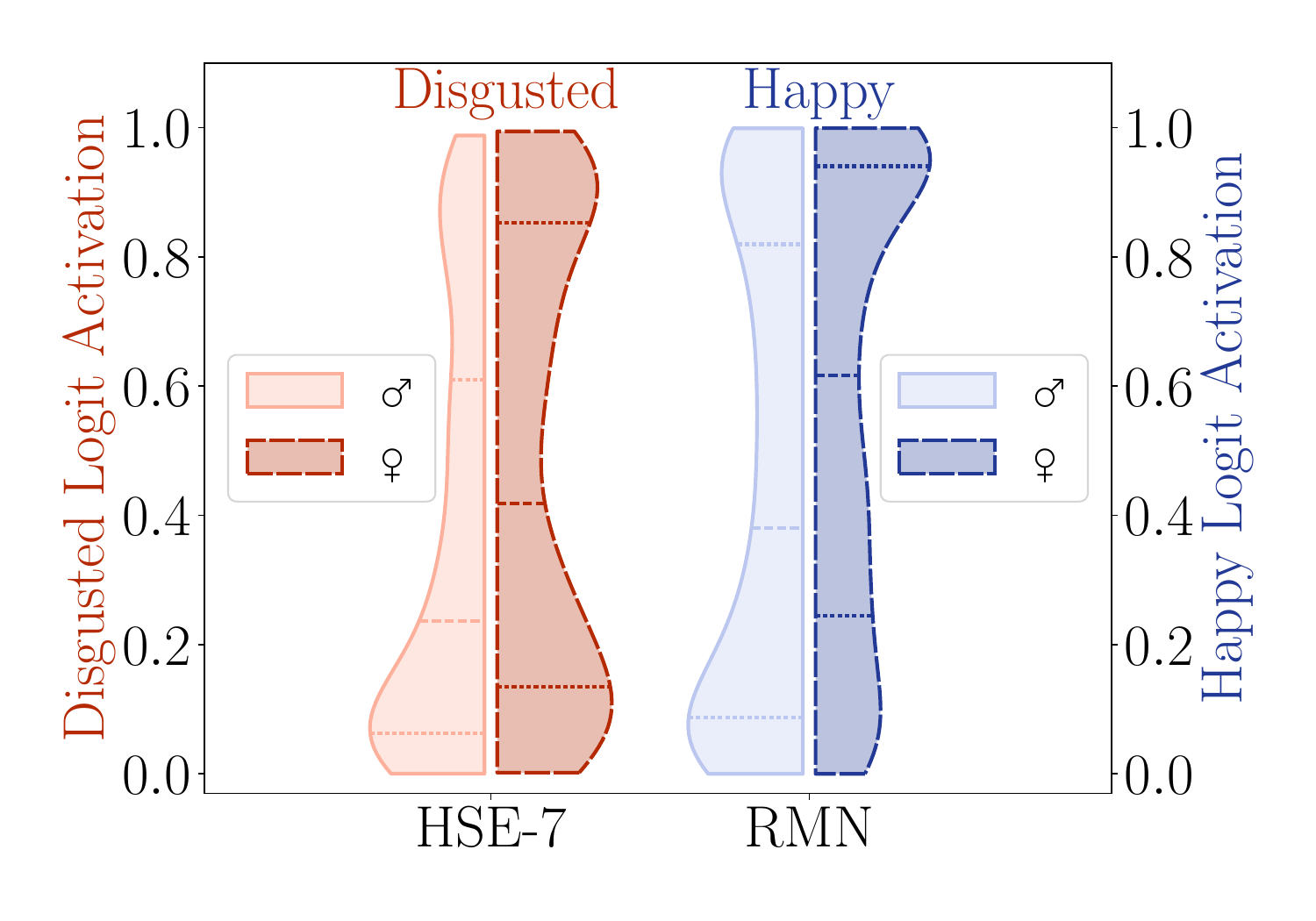}
        \caption{Gender ($\bod_G$)}
        \label{fig:analysis_gender}
    \end{subfigure}
    \begin{subfigure}{0.64\textwidth}
        \centering
        \includegraphics[width=\textwidth, clip, trim={0cm 1.41cm 0cm 1.1cm}]{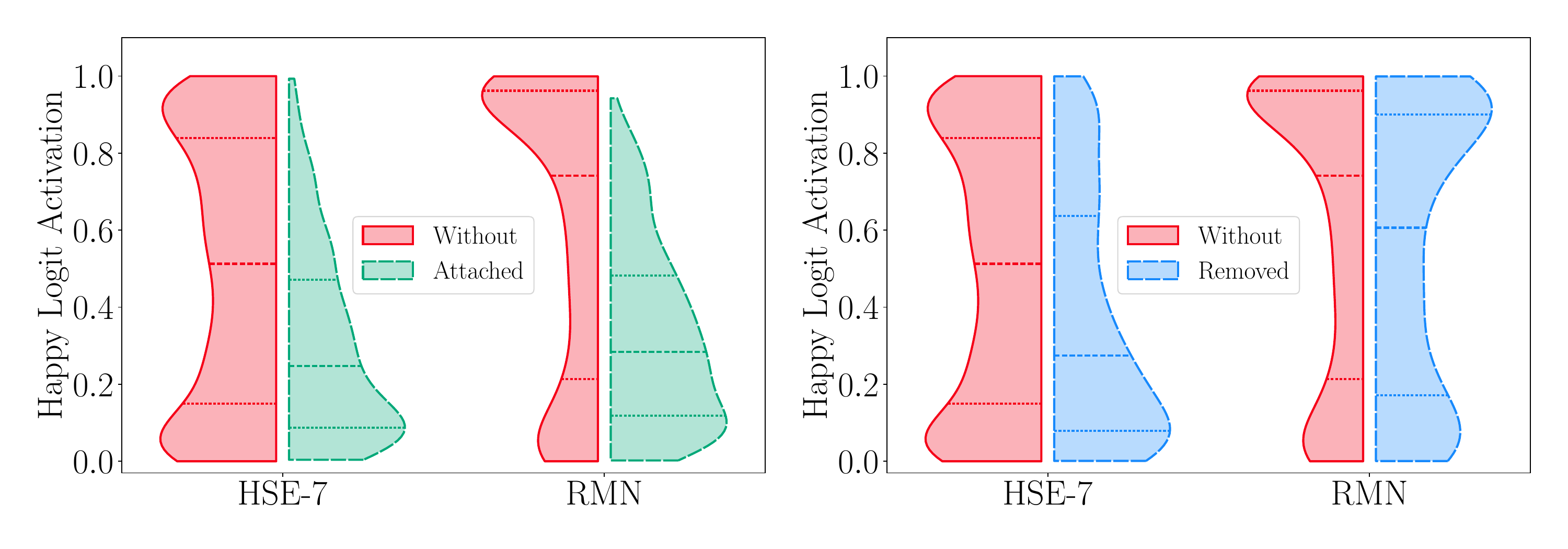}
        \caption{Attached sEMG ($\rec_A$) and removed sEMG ($\rec_R$)}
        \label{fig:analysis_emg}
    \end{subfigure}
    \begin{subfigure}{0.32\textwidth}
        \centering
        \includegraphics[width=\textwidth, clip, trim={0cm 1.41cm 0cm 1.1cm}]{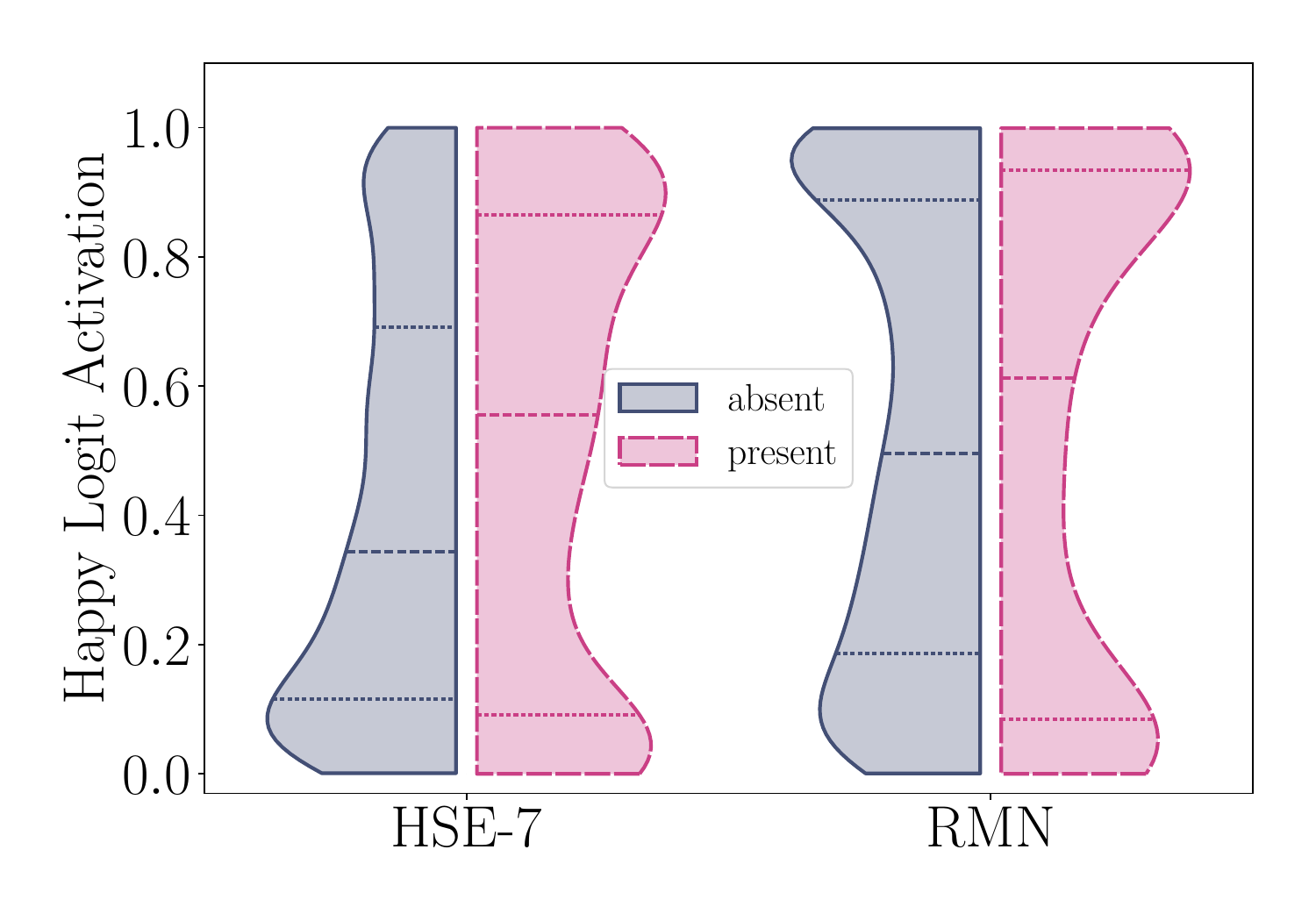}
        \caption{Facial palsy ($\bod_F$)}
        \label{fig:analysis_fcp}
    \end{subfigure}
    \begin{subfigure}{0.64\textwidth}
        \centering
        \includegraphics[width=\textwidth, clip, trim={0cm 1.41cm 0cm 1.1cm}]{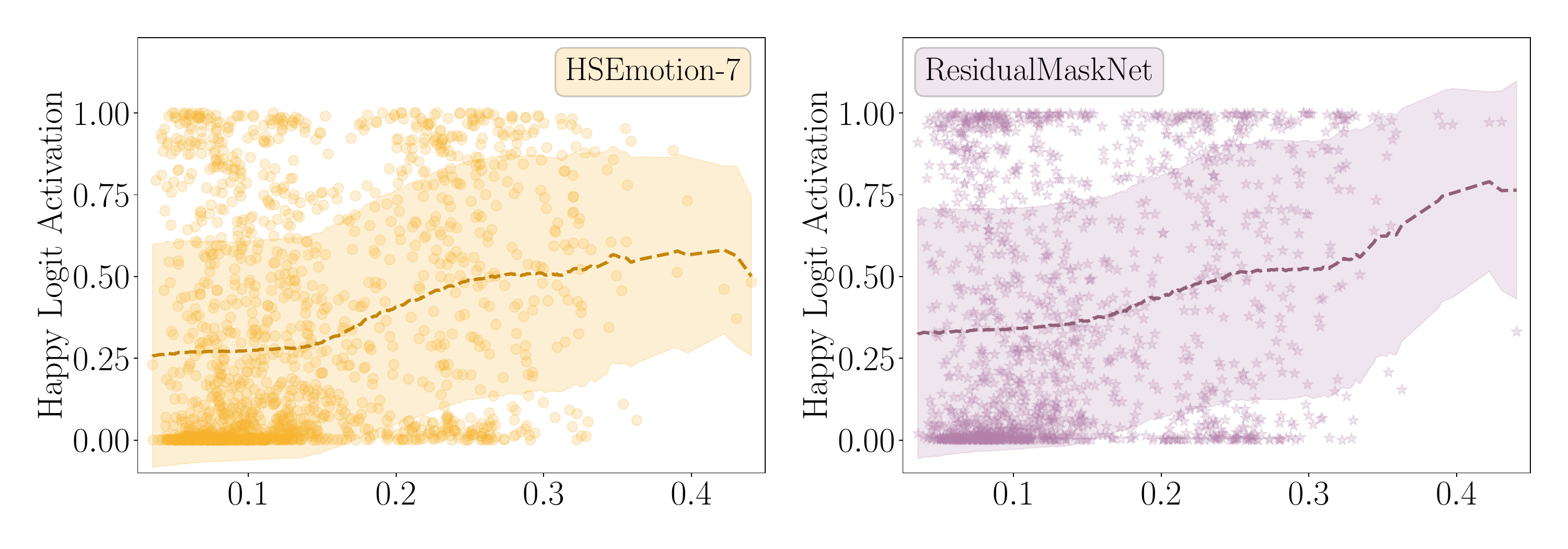}
        \caption{LPIPS ($\sym_L$) symmetry value for lateral face halves}
        \label{fig:analysis_lpips}
    \end{subfigure}\\
    \caption{
        Model behavior visualization:
        In the binary \binary~case, we use violin plots for logit distributions (\subref{fig:analysis_gender}, \subref{fig:analysis_emg}, \subref{fig:analysis_fcp}).
        For the scalar \scalar~\emph{property} $\sym_L$, we plot a regression analysis by estimating Gaussians with a sliding window~\cite{penzel2023analyzing}.
        For all visualizations, we only use images of the corresponding emotion~\cite{reimers2020determining}.
    }
    \label{fig:behavior_analysis}
\end{figure}

We investigate subpopulation shifts resulting from different \emph{property} manifestations by applying black-box neural networks.
Specifically, we focus on ResidualMaskNet (RMN)~\cite{pham2021facial} and HSEmotion-7 (HSE-7)~\cite{savchenko2023facial} for facial emotion recognition.
Each image of our dataset has all \emph{properties}, as referenced in \autoref{tab:features}.
Using these images, we capture the logit activations of the pre-trained RMN and HSE-7.
Hence, using the reference annotations, we test for significant changes in behavior~\cite{reimers2020determining}. 
Statistically significant results ($p < 0.01$) are denoted per model and emotion in \autoref{tab:ci-results}.
Of the $84$ possible combinations, RMN utilizes information in $91.25\%$ (\nicefrac{73}{84}) and for HSE-7 in $87.50\%$ (\nicefrac{70}{84}) of cases. 
The findings show that the models use \emph{properties} in their decision-making processes.
We now examine the models' behavior concerning several \emph{properties} in detail.

First, we focus on self-declared gender.
Other work~\cite{chenUnderstandingMitigatingAnnotation2021} demonstrated that gender ($\bod_G$) affects emotion classification, a claim we corroborate for $11$ of $14$ cases as seen in \autoref{tab:ci-results}.
The visualization, see \autoref{fig:analysis_gender}, displays the logit activation distribution regarding men and women.
We choose \emph{disgusted} for HSE-7 due to its significance and difficulty in execution~\cite{guntinas2023high,buechner2023improved,Buechner20232D3DFace}; see \autoref{tab:ci-results}.
The remaining investigations focus on \emph{happy} expressions due to the observed prediction stability for probands and patients; see \autoref{tab:ci-results}.
The displayed distributions indicate HSE-7 struggles to classify \emph{disgusted} expressions for men compared to women.
Moreover, the RMN more frequently attributes \emph{happy} expressions to women.

Secondly, we investigate the attached ($\rec_E$) and artificially removed ($\rec_R$) sEMG electrodes, as they are influential for both models and all seven emotions; see \autoref{tab:ci-results}.
As argued above, we visualize the \emph{happy} logit distribution concerning $\rec_E$ and $\rec_R$ in \autoref{fig:analysis_emg}.
The overall lower activations are anticipated for the attached case~\cite{buechner2023lets,buechner2023improved}.
Yet we see in \autoref{tab:accuracy} that the quality of the removal highly depends on the model and emotion when recovering predicted performance.
For instance, regarding \emph{happy} expressions, the RMN recovers a higher predictive performance and logit distribution (see \autoref{fig:analysis_emg}).
We assume that the recovery introduced artifacts or the sEMG influences the participants' facial mimicry.

Lastly, we assess whether facial palsy-induced asymmetry ($\bod_F$) or symmetry in general ($\sym$) impacts model behavior.
In \autoref{tab:ci-results}, we observe in $77.14\%$ (\nicefrac{54}{70}) of combinations a significant influence of facial symmetry.
Especially, facial palsy ($\bod_F$) and different LPIPS ($\sym_L$) manifestations influence the behavior of both models and all seven emotions.
We follow \cite{penzel2023analyzing} and regress the \emph{property} manifestation trend by estimating Gaussians with a sliding window approach for LPIPS.
Again, focusing on \emph{happy} expressions, \autoref{fig:analysis_lpips} show that higher facial symmetry leads to higher logit activation on average.
However, facial palsy specifically, see \autoref{fig:analysis_fcp}, shows increased uncertainty.
The indicated quartiles suggest a strong bimodal distribution with many high and low activations.
This suggests that the models are influenced by facial symmetry.
Hence, their application for unilateral facial palsy patients should be approached with caution.

\section{Conclusion}
This work studied emotion classifiers applied in a medical context.
To go beyond performance metrics, we used the causal-based framework from \cite{reimers2020determining}.
We demonstrated that up to $91.25\%$ of classifier output behavior changes are statistically significant concerning varying \emph{properties}, including age, gender, and facial symmetry.
To obtain such \emph{properties} unaccounted in other datasets, we recorded 36 probands and 36 patients with facial palsy, a disease affecting facial expressions.

To summarize, we observe differences in model behavior regarding gender and facial symmetry.
Hence, their application on medical conditions should be approached with care.
Additionally, the obstruction of facial features during medical studies can significantly impact the model behavior, as we observe for attached sEMG electrodes.
However, our observations do not necessarily indicate harmful biases but open the discussion beyond simple predictive performance.

Finally, our selected \emph{properties} are not exhaustive and do not cover all possible biases.
Many other properties related to different downstream tasks could be conceived and studied in the future.
Additionally, different models and \emph{property} aware training are promising~\cite{blunk2023beyond,chenUnderstandingMitigatingAnnotation2021}.
We hope to prompt more extensive analysis and inspire researchers to analyze facial recognition models beyond prediction performance.

\begin{credits}
\subsubsection{\ackname} Partially supported by Deutsche Forschungsgemeinschaft (DFG - German Research Foundation) project 427899908 BRIDGING THE GAP: MIMICS AND MUSCLES (DE 735/15-1 and GU 463/12-1).

\subsubsection{\discintname}
The authors have no competing interests to declare.
\end{credits}

\bibliographystyle{splncs04}
\bibliography{sources}

\begin{thebibliography}{10}
\providecommand{\url}[1]{\texttt{#1}}
\providecommand{\urlprefix}{URL }
\providecommand{\doi}[1]{https://doi.org/#1}

\bibitem{blunk2023beyond}
Blunk, J., Penzel, N., Bodesheim, P., Denzler, J.: Beyond debiasing: Actively
  steering feature selection via loss regularization. In: DAGM German
  Conference on Pattern Recognition (DAGM-GCPR) (2023)

\bibitem{buechner2023improved}
B{\"u}chner, T., Guntinas-Lichius, O., Denzler, J.: Improved obstructed facial
  feature reconstruction for emotion recognition with minimal change cyclegans.
  In: Advanced Concepts for Intelligent Vision Systems (Acivs). pp. 262--274.
  SpringerNature (august 2023). \doi{10.1007/978-3-031-45382-3_22}

\bibitem{buechner2023lets}
B{\"u}chner, T., Sickert, S., Volk, G.F., Anders, C., Guntinas-Lichius, O.,
  Denzler, J.: Let’s get the facs straight - reconstructing obstructed facial
  features. In: International Conference on Computer Vision Theory and
  Applications (VISAPP). SciTePress (march 2023).
  \doi{10.5220/0011619900003417}

\bibitem{buchnerFacesVolumesMeasuring2023}
B{\"u}chner, T., Sickert, S., Volk, G.F., {Guntinas-Lichius}, O., Denzler, J.:
  From {{Faces}} to~{{Volumes}} - {{Measuring Volumetric Asymmetry}} in~{{3D
  Facial Palsy Scans}}. In: Advances in {{Visual Computing}}. Lecture {{Notes}}
  in {{Computer Science}}, {Springer Nature Switzerland} (2023).
  \doi{10.1007/978-3-031-47969-4_10}

\bibitem{Buechner20232D3DFace}
Büchner, T., Sickert, S., Graßme, R., Anders, C., Guntinas-Lichius, O.,
  Denzler, J.: Using 2d and 3d face representations to generate comprehensive
  facial electromyography intensity maps. In: International Symposium on Visual
  Computing (ISVC). pp. 136--147 (2023). \doi{10.1007/978-3-031-47966-3_11},
  \url{https://link.springer.com/chapter/10.1007/978-3-031-47966-3_11}

\bibitem{chenUnderstandingMitigatingAnnotation2021}
Chen, Y., Joo, J.: Understanding and {{Mitigating Annotation Bias}} in {{Facial
  Expression Recognition}}. In: 2021 {{IEEE}}/{{CVF International Conference}}
  on {{Computer Vision}} ({{ICCV}}). pp. 14960--14971. {IEEE}, {Montreal, QC,
  Canada} (Oct 2021). \doi{10.1109/ICCV48922.2021.01471}

\bibitem{ekman1992argument}
Ekman, P.: An argument for basic emotions. Cognition and Emotion
  \textbf{6}(3-4),  169--200 (1992). \doi{10.1080/02699939208411068}

\bibitem{fukumizu2007kernel}
Fukumizu, K., Gretton, A., Sun, X., Sch{\"o}lkopf, B.: Kernel measures of
  conditional dependence. Advances in neural information processing systems
  \textbf{20} (2007)

\bibitem{guntinas2023high}
Guntinas-Lichius, O., Trentzsch, V., Mueller, N., Heinrich, M., Kuttenreich,
  A.M., Dobel, C., et~al.: High-resolution surface electromyographic activities
  of facial muscles during the six basic emotional expressions in healthy
  adults: a prospective observational study. Scientific Reports
  \textbf{13}(1),  19214 (2023)

\bibitem{pearl2009causality}
Pearl, J.: Causality. Cambridge university press (2009)

\bibitem{penzel2023analyzing}
Penzel, N., Kierdorf, J., Roscher, R., Denzler, J.: Analyzing the behavior of
  cauliflower harvest-readiness models by investigating feature relevances. In:
  2023 IEEE/CVF International Conference on Computer Vision Workshops (ICCVW).
  pp. 572--581. IEEE (2023)

\bibitem{penzel2022investigating}
Penzel, N., Reimers, C., Bodesheim, P., Denzler, J.: Investigating neural
  network training on a feature level using conditional independence. In:
  European Conference on Computer Vision. pp. 383--399. Springer (2022)

\bibitem{pham2021facial}
Pham, L., Vu, T.H., Tran, T.A.: Facial expression recognition using residual
  masking network. In: 2020 25th International Conference on Pattern
  Recognition (ICPR). pp. 4513--4519 (2021).
  \doi{10.1109/ICPR48806.2021.9411919}

\bibitem{reichenbach1956direction}
Reichenbach, H.: The direction of time, vol.~65. Univ of California Press
  (1956)

\bibitem{reimers2021conditional}
Reimers, C., Penzel, N., Bodesheim, P., Runge, J., Denzler, J.: Conditional
  dependence tests reveal the usage of abcd rule features and bias variables in
  automatic skin lesion classification. In: Proceedings of the IEEE/CVF
  Conference on Computer Vision and Pattern Recognition. pp. 1810--1819 (2021)

\bibitem{reimers2020determining}
Reimers, C., Runge, J., Denzler, J.: Determining the relevance of features for
  deep neural networks. In: European Conference on Computer Vision. Springer
  (2020)

\bibitem{runge2018conditional}
Runge, J.: Conditional independence testing based on a nearest-neighbor
  estimator of conditional mutual information. In: International Conference on
  Artificial Intelligence and Statistics. PMLR (2018)

\bibitem{savchenko2023facial}
Savchenko, A.: Facial expression recognition with adaptive frame rate based on
  multiple testing correction. In: International Conference on Machine
  Learning. vol.~202. PMLR (2023),
  \url{https://proceedings.mlr.press/v202/savchenko23a.html}

\bibitem{shah2020hardness}
Shah, R.D., Peters, J.: The hardness of conditional independence testing and
  the generalised covariance measure. The Annals of Statistics  \textbf{48}(3),
   1514--1538 (2020)

\bibitem{strobl2019approximate}
Strobl, E.V., Zhang, K., Visweswaran, S.: Approximate kernel-based conditional
  independence tests for fast non-parametric causal discovery. Journal of
  Causal Inference  (2019)

\bibitem{weiAssessingFacialSymmetry2022}
Wei, W., Ho, E.S.L., McCay, K.D., Dama{\v s}evi{\v c}ius, R., Maskeli{\=u}nas,
  R., Esposito, A.: Assessing {{Facial Symmetry}} and {{Attractiveness}} using
  {{Augmented Reality}}. Pattern Analysis and Applications  \textbf{25}(3)
  (2022). \doi{10.1007/s10044-021-00975-z}

\bibitem{yang2023change}
Yang, Y., Zhang, H., Katabi, D., Ghassemi, M.: Change is hard: A closer look at
  subpopulation shift. arXiv preprint arXiv:2302.12254  (2023)

\bibitem{zhangUnreasonableEffectivenessDeep2018}
Zhang, R., Isola, P., Efros, A.A., Shechtman, E., Wang, O.: The {{Unreasonable
  Effectiveness}} of {{Deep Features}} as a {{Perceptual Metric}}. Proceedings
  of the IEEE Conference on Computer Vision and Pattern Recognition  (Apr
  2018). \doi{10.48550/arXiv.1801.03924}

\end{thebibliography}

\end{document}